\documentclass[10pt,letterpaper]{article}
\usepackage[top=0.85in,left=2.25in,footskip=0.75in,marginparwidth=2in]{geometry}

\usepackage[utf8]{inputenc}

\usepackage{cite}

\usepackage{nameref,hyperref}

\usepackage[right]{lineno}

\usepackage{microtype}
\DisableLigatures[f]{encoding = *, family = * }

\usepackage{changepage}

\usepackage[aboveskip=1pt,labelfont=bf,labelsep=period,singlelinecheck=off]{caption}

\makeatletter
\renewcommand{\@biblabel}[1]{\quad#1.}
\makeatother

\usepackage{lastpage,fancyhdr,graphicx}
\usepackage{tikz}
\usepackage{amsmath}
\usepackage{amsfonts}
\usepackage{booktabs}
\usepackage{subcaption}
\usepackage{float}
\usepackage{epstopdf}
\fancyheadoffset[L]{1.75in}
\fancyfootoffset[L]{1.75in}

\usepackage{color}

\definecolor{Gray}{gray}{.25}

\usepackage{graphicx}

\usepackage{sidecap}

\usepackage{wrapfig}
\usepackage[pscoord]{eso-pic}
\usepackage[fulladjust]{marginnote}
\reversemarginpar

\newcommand{\orcidicon}{\includegraphics[width=0.32cm]{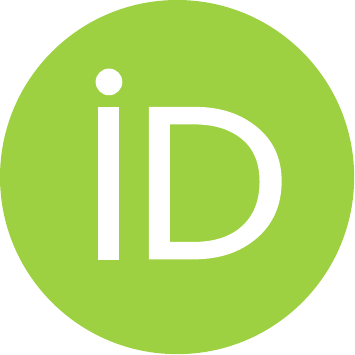}}
\foreach \x in {A, ..., Z}{%
\expandafter\xdef\csname orcid\x\endcsname{\noexpand\href{https://orcid.org/\csname orcidauthor\x\endcsname}{\noexpand\orcidicon}}
}

\begin{document}
\vspace*{0.35in}

\begin{flushleft}
{\Huge
\textbf\newline{Eigenbehaviour as an Indicator of Cognitive Abilities}
}
\newline

\newcommand{\orcidauthorA}{0000-0002-1245-2931} 
\newcommand{\orcidauthorB}{0000-0001-5069-407X} 
\newcommand{\orcidauthorC}{0000-0003-0625-360X} 
\newcommand{\orcidauthorD}{0000-0003-1584-0312} 
\newcommand{\orcidauthorE}{0000-0001-6990-4188} 
\newcommand{\orcidauthorF}{0000-0002-8069-9450} 
\newcommand{\orcidauthorG}{0000-0002-2425-0077} 

Angela Botros\textsuperscript{1}\orcidA{},
Narayan Schütz\textsuperscript{1}\orcidB{}{},
Christina Röcke\textsuperscript{1}\orcidC{},
Robert Weibel\textsuperscript{1}\orcidG{},
Mike Martin\textsuperscript{1}\orcidD{},
René Müri\textsuperscript{1}\orcidE{},
Tobias Nef\textsuperscript{1,*}\orcidF{}
\\
\bigskip
\bf{1} ARTORG Center for Biomedical Engineering Research, Gerontechnology and Rehabilitation, University of Bern, Bern, Switzerland\\
\bf{2} University Research Priority Program (URPP) "Dynamics of Healthy Aging", University of Zurich, Andreasstrasse, Zurich, Switzerland \\
\bf{3} Department of Geography, University of Zurich, Zurich, Switzerland \\
\bf{4} Department of Psychology, University of Zurich, Binzmühlestrasse, Zurich, Switzerland \\
\bf{5} Department of Neurology, Inselspital, University Hospital of Bern, University of Bern, Switzerland \\
* tobias.nef@unibe.ch

\end{flushleft}

\section*{Abstract}
With growing usage of machine learning algorithms and big data in health applications, digital biomarkers have become an important key feature to ensure the success of those applications.
In this paper, we focus on one important use-case, the long-term continuous monitoring of the cognitive ability of older adults.
The cognitive ability is a factor both for long-term monitoring of people living alone as well as an outcome in clinical studies.
In this work, we propose a new digital biomarker for cognitive abilities based on location eigenbehaviour obtained from contactless ambient sensors.
Indoor location information obtained from passive infrared sensors is used to build a location matrix covering several weeks of measurement.
Based on the eigenvectors of this matrix, the reconstruction error is calculated for various numbers of used eigenvectors.
The reconstruction error is used to predict cognitive ability scores collected at baseline, using linear regression.
Additionally, classification of normal versus pathological cognition level is performed using a support-vector-machine.
Prediction performance is strong for high levels of cognitive ability, but grows weaker for low levels of cognitive ability. 
Classification into normal versus pathological cognitive ability level reaches high accuracy with a AUC = 0.94.
Due to the unobtrusive method of measurement based on contactless ambient sensors, this digital biomarker of cognitive ability is easily obtainable.
The usage of the reconstruction error is a strong digital biomarker for the binary classification and, to a lesser extent, for more detailed prediction of interindividual differences in cognition.

\section*{Introduction}
The monitoring of cognitive ability of people is a task to assess the health state of adults.
This holds true for both conducting clinical trials~\cite{Craft_Intranasal_2012, Valls_Mediterranean_2015, forlenza_radanovic_talib_gattaz_2019}, as well as home-monitoring for elderly people \cite{NGANDU20152255, Seelye_Feasibility_2020}.
In both cases, continuous long-term monitoring can complement the existing point-in-time examinations, done by medical professionals in a clinic~\cite{mckay_long-term_2019}.

For the measurement of cognitive abilities, there are a range of different tests.
The Mini-Mental State Exam (MMSE) is an assessment to test for mild cognitive impairment (MCI), Alzheimer's disease (AD) and other cognitive issues~\cite{folstein1975mini}.
This test has a maximum cognitive ability score of 30.
A cutoff is used to distinguish between normal cognitive ability and people with dementia.
A generally accepted cutoff score for the MMSE is 24~\cite{folstein1975mini}, but a more differentiated interpretation is suggested with adaptive cutoffs depending of the peoples age and education~\cite{iverson_interpretation_1998}.
Another test to detect cognitive impairment is the Montreal Cognitive Assessment (MoCA)~\cite{nasreddine_montreal_2005}.
This test has a maximum score of 30, too.
The cutoff to distinguish between people with normal cognitive ability and people with MCI or AD is set at 26~\cite{ihle-hansen_montreal_2017}; but this value has been up for debate~\cite{nasreddine_montreal_2005}.
Other evaluations usually cover a selection of individual tasks, such as the Halstead-Reitan neuropsychological test battery~\cite{reitan1986halstead}.

Pervasive computing is a methodology that is promising for long-term monitoring of cognitive abilities.
It could provide safety and security for independently living individuals, while still maintaining simplicity and - depending on the exact chosen technology - privacy.
Several studies have provided evidence for both acceptance by the people and usefulness of such systems \cite{stucki_web-based_2014, schutz_validity_2019, lyons_pervasive_2015, alam_automated_2016, Garcia_Ambient_2020, urwyler_cognitive_2017}.

Stucki et al. have developed and tested a sensor system based on ambient and object sensors to track activities of daily living (ADL)~\cite{stucki_web-based_2014}.
They have shown that this system is able to track ADL reliably, both in healthy individuals and people with AD.
Schütz et al. were able to track physical health reliably over a year, using a pervasive sensor system~\cite{schutz_validity_2019}.
Their system consisted of ambient, object and wearable sensors and was used for the monitoring of older adults, living alone.
Another multi faceted system consisting of ambient and object sensors was used by Lyons et al. to monitor older people~\cite{lyons_pervasive_2015}.
Over the span of eight years they monitored their participants with the goal to continually asses the cognitive status.
Alam et al. have used wearable sensors to track physiological data in order to 
assess cognitive ability trajectories in the context of dementia~\cite{alam_automated_2016}.

With regards to the cognitive ability, Urwyler et al. have shown that the daily routine of a a person diagnosed with Alzheimer's dementia is far less regular and more chaotic than the daily behaviour of age-matched healthy individuals~\cite{urwyler_cognitive_2017}.
They have monitored people living alone using passive infrared (PIR) sensors, a type of sensor detecting movement while maintaining maximal privacy.
Based on the results of Urwyler et al., we hypothesise, that the regularity of people's movement patterns at home is correlated to their cognitive ability and thus influences the eigenvectors constructed from their behaviour matrix.

The idea that certain medical conditions can affect daily behaviour has been studies by Paraschiv-Ionescu et al.m where they monitored physical activities of chronic pain patients over multiple days, and compared the data to those of healthy pain-free people. They specifically noted the change in complexity of physical activities between chronic-pain patients and pain-free patients, with the latter showing a more complex behaviour~\cite{paraschiv-ionescu_barcoding_2012}.

For the analysis of the behaviour regularity we suggest a method based on eigen-decomposition of behavioural matrices, a method introduced by Eagle et al.~\cite{eagle_eigenbehaviors:_2009}.
They used the approximate localisation data obtained from cell phones of 100 participants.
An eigenvalue decomposition on this data provided insight into the students' behaviour, organisational group and circle of friends.
The idea to use principal component analysis to analyse the underlying structure of data is not new, it was notably used in~\cite{sirovich_low-dimensional_1987}, where principal component analysis is conducted to represent faces.

In the current work, we introduce a method to assess older adults' cognitive ability, based on movement patterns obtained through unobtrusive ambient sensor technology.
We evaluate this method based on data obtained from 48 individuals above the age of retirement.

\section*{Materials and Methods}
\subsection*{Participants}
The data presented in this study stems from two studies, the StrongAge Cohort Study and the MOASIS MobiPro Study. Both studies were conducted based on the principles declared in the Declaration of Helsinki and approved by the University Ethics Committee.
All participants signed and handed in an informed consent before study participation.\newline
The \textit{StrongAge Cohort Study} is a home-monitoring study, where community dwelling seniors (inclusion criterion $\geq 80$ years) were equipped with pervasive computing systems for approximately one year~\cite{schutz_validity_2019}.
The recruitment aimed to represent a naturalistic sample of alone living older adults in central Switzerland, irrespective of their cognitive ability.\newline
The \textit{MOASIS MobiPro Study} is a home-monitoring study, where community dwelling seniors (inclusion criterion $\geq 65$ years, MMSE score $\geq 27$) were equipped with pervasive computing systems for approximately four weeks~\cite{MobiPro_2020}.
The aim of the study was to assess mobility, physical and social acitivity patterns in relation to health and well-being in healthy older adults.
For the current analysis, only alone living participants of the MOASIS MobiPro Study were selected.

\subsection*{Data collection}
In this study, passive infrared (PIR) sensors (DomoSafety SA) were used to monitor the participants in their respective homes.
The PIR sensors were placed in order to cover the relevant living spaces: bedroom, kitchen, bathroom, living room and entrance area. 
These sensors recorded presence or absence of movement with a frequency of $0.5 Hz$.
In addition to the PIR sensors, door sensors were placed on the entrance door and the fridge to assess time outside of home as well as kitchen usage.
The sensors were installed in the participants' homes at the beginning of the study and disassembled again at the end.
At the beginning of the respective studies, participants' cognitive ability was assessed with a diverse battery of tests, including the MMSE.
The MMSE score was used in this project as measure of cognitive ability.
The sensor based activity and mobility monitoring in the StrongAge Cohort Study was done for up to a year.
The monitoring in the MOASIS study was over a span of four weeks.
To avoid any biases, the data from the StrongAge Cohort Study was sub-sampled.
Both time-points of measurement as well as number of days distributions were matched.
The obtained data is not publicly available due to local Swiss data regulations. 

\subsection*{Behaviour Matrix and Eigendecomposition}
\begin{figure}[H]
\centering
\begin{subfigure}[t]{0.45\textwidth}
    \centering
    \includegraphics[width=0.9\textwidth]{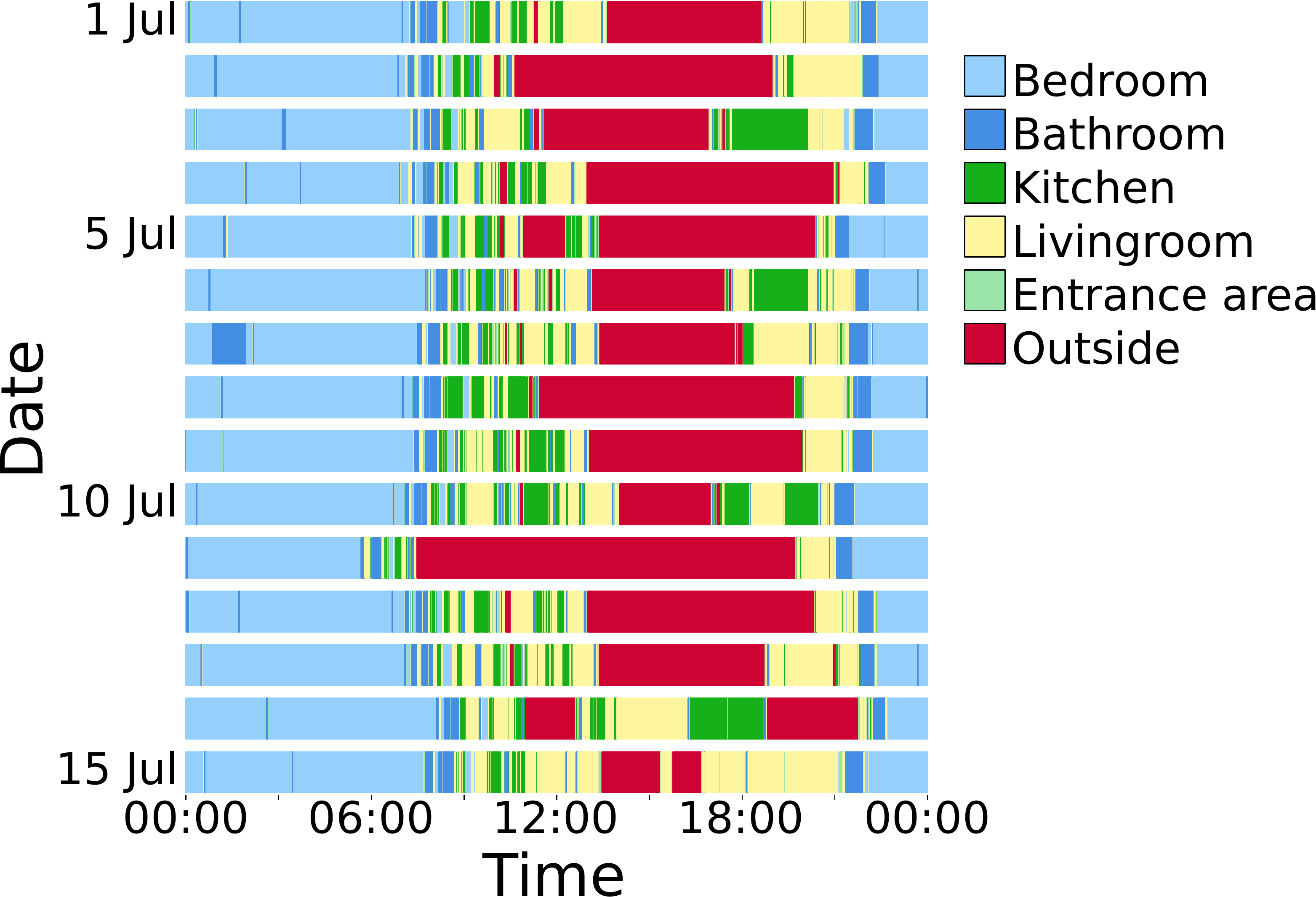}
    \caption{}
    \label{fig:ColourLocationMatrix}
\end{subfigure}
\begin{subfigure}[t]{0.45\textwidth}
    \centering
    \includegraphics[width=0.9\textwidth]{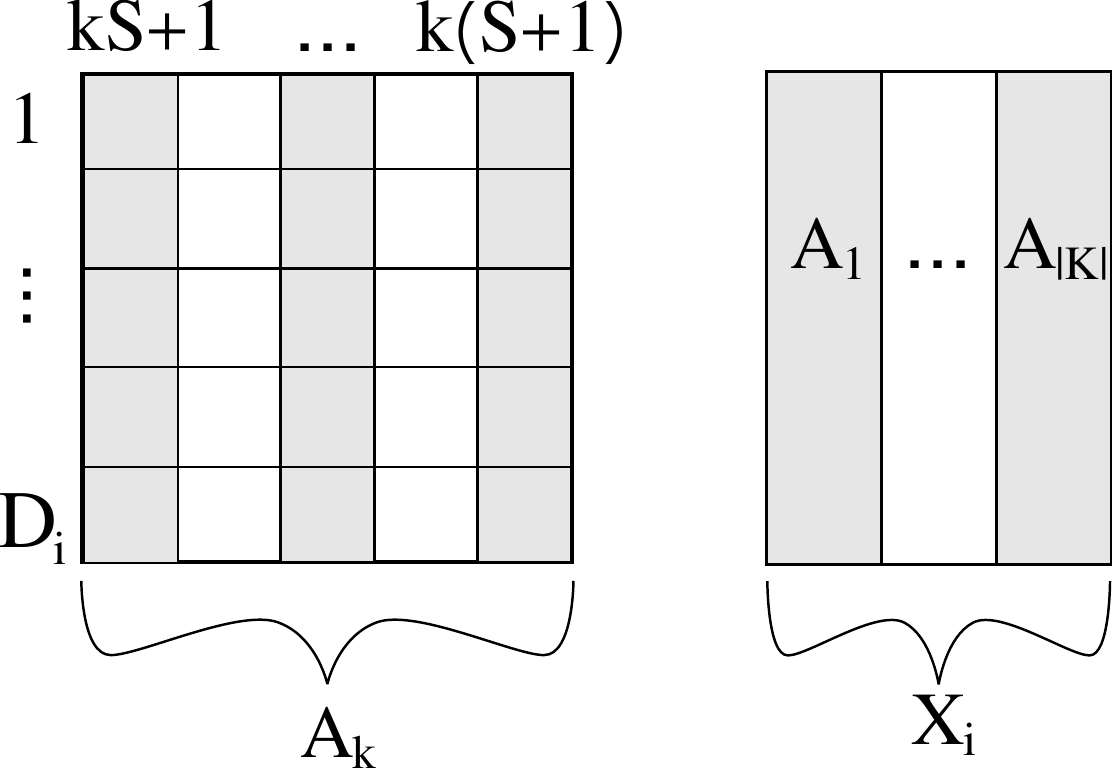}
    \caption{}
    \label{fig:LocationMatrix}
\end{subfigure}%
\label{fig:ExplanationIntro}
\caption{
In \textbf{a)} is shown a representation of multiple days, with different room-locations colour-coded. 
In \textbf{b)}, the structure of matrix $X^i$ is explained. 
Along the columns, sub-matrices $A_k$ are stacked.
Every sub-matrix $A_k$ contains the presence percentages of location $k$.
Its width is $S$ and its height is $D_i$, the total measurement days of person $i$.
The total matrix $X^i$ consists of the horizontally stacked matrices $A_k$.
}
\end{figure}

The PIR sensor data consists of time and duration of activation for all sensors.
Based on this, the location of the people in their apartment throughout each day was obtained.
The set of locations is $K = \{bedroom,~ bathroom,~ living room,~ kitchen, ~entrance, ~outside\}$.
A visual representation of the locations as estimated by the sensors is given in Figure~\ref{fig:ColourLocationMatrix}.
For the eigenbehaviour, every day of data is subdivided into $S$ time windows, each of length $\Delta t = \frac{24h}{S}$. 
Different number of $S$ were assessed, with $S:= \{24, 48, 96, 144, 288\}$ resulting in window lengths from 5 minutes $(S=288)$ up to one hour $(S=24)$.
For every time window, the percentage of presence in every room was calculated.

For every person $i$, a location matrix $X^i$ was computed.
Every row is a day of measurement, with a total height of $D^i$-measurement days for person $i$.
In the columns, the percentages of presence for every time window and location are given.
The locations are stacked horizontally, \textit{i.e.} the first $S$ columns represent the first location, and the columns $\{(k-1)\cdot S, (k-1)\cdot S + 1, \hdots, k \cdot S - 1\}$ represent the time windows of location $K$ for $k = 1 \hdots |K|$.
The resulting $X^i$ is a $|D^i| \times S\cdot |K|$ matrix.
This is also shown in Figure~\ref{fig:LocationMatrix}.

In every individual cell $X^i[d_j, k\cdot S +n]$, the percentage of presence in the corresponding location $K$ on day $d_j$ in the time window $[\frac{24h}{\Delta t}n, \frac{24h}{\Delta t}(n+1)]$ is given.
This fragmentation was the same for all people.
The $j-th$ row of $X^i$ is $\Gamma^i_j$ and represents exactly one day, or one point in an ($S\cdot |K|$)-dimensional space.
The average location-vector of person $i$ is $\Psi^i = \frac{1}{D^i}\sum_{d=1}^{D^i}\Gamma^i_j$. 
The deviation of an individual day from the average day is $\Phi^i_j = \Gamma^i_j - \Psi^i$.
The location deviation matrix is $\hat{X}^i = [\Phi^i_1,\cdots, \Phi^i_{D^i}] \in \mathbb{R}^{D^i\times (S\cdot |K|)}$.
To analyse the different behaviours for every person $i$, principal component analysis is performed on the collection of vectors $\Phi^i_j$.
The covariance matrix $C^i$ of person $i$ is based on this set:
\begin{equation}
    C^i = \hat{X}^i(\hat{X}^i)^T,  ~~\textrm{s.t.}~~ c^i[n,m] = \sum_{d=1}^{D^i}\Phi^i_d[n] \Phi^i_d[m]
\end{equation}
From the covariance matrix $C^i$ of person $i$, the Eigenvectors $v^i_l$ and Eigenvalues $\lambda^i_l(C^i)$ can be computed.
They represent the principal components of the deviation vectors $\Phi^i_j$. \newline
Based on the set of Eigenvectors $v^i_l$ and Eigenvalues $\lambda^i_l(C^i)$ of person $i$, the matrix $\hat{X}^i$ can be reconstructed again.
Depending on the number of Eigenvectors and Eigenvalues that are used for the reconstruction, the resulting reconstructed matrix will deviate from the original one.
The difference between the reconstructed matrix and the original matrix is the reconstruction error.
We will refer to the reconstruction error obtained from using only the first eigenvector as the first reconstruction error, and the reconstruction error obtained from using the first $n$ eigenvectors as the $n-th$ reconstruction error.
Due to a differing number of measurement days $D^i$, the matrices $\hat{X}^i$ have varying size.
To normalize the reconstruction error, the sum of the absolute deviations is divided by size of the matrix $D^i\cdot S \cdot |K|$.
This gives a mean deviation per matrix segment.

\subsection*{Prediction and classification of cognitive ability}
In order to assess the influence of age on the prediction of the cognition score, the partial correlation of the cognition score and the reconstruction error was computed, with  age as the other confounding variable.

The cognition score was predicted with a linear regression, using the reconstruction error and age as features.
As a baseline, a regression was trained using only age, no reconstruction error.
Cross-validation is used for evaluation, due to the small sample size.
The root-mean-squared deviation $RMSD = \frac{1}{N}\sum_i((y_i - \hat{y}_i)^2)$ of the predicted score $\hat{y}_i$ and true score $y_i$  over all $N$ measured people describes the measurement of accuracy.

Besides the cognition-prediction, a more general classification was performed, where the participants were divided into two groups; those with a score at or above 26, and those with a score below 26.
As a reference, a score at or above 26 is considered normal, while a score below this value indicates mild to severe cognitive impairment~\cite{nasreddine_montreal_2005}.
For the cross-validation, the split was done with two-thirds of the data (= 32 samples) in the training set and one third (=16 samples) in the validation set, with the two sets stratified.
The receiver-operating-characteristic (ROC) and its area under the curve (AUC) were calculated for all cross-validation folds.
The mean ROC and the confidence interval ($\pm std)$ were evaluated for final assessment.

The optimization parameters were the number of eigenvectors used for the reconstruction and the size of the time window $S$.
For both prediction and classification, a simultaneous grid-search was performed in a leave-one-out evaluation for parameter optimization.

All preprocessing and calculations were done using the Python programming language version 3.6.9 (Python Software Foundation). 
Correlations and significances thereof were calculated using the Python package scipy.stats, version 1.3.1.
Figures and graphical illustration were created using the above mentioned Python programming language, as well as Inkscape version 1.0.








\section*{Results}
\begin{figure}[H]
\centering
\begin{subfigure}[t]{0.4\textwidth}
    \centering
    \includegraphics[width=\textwidth]{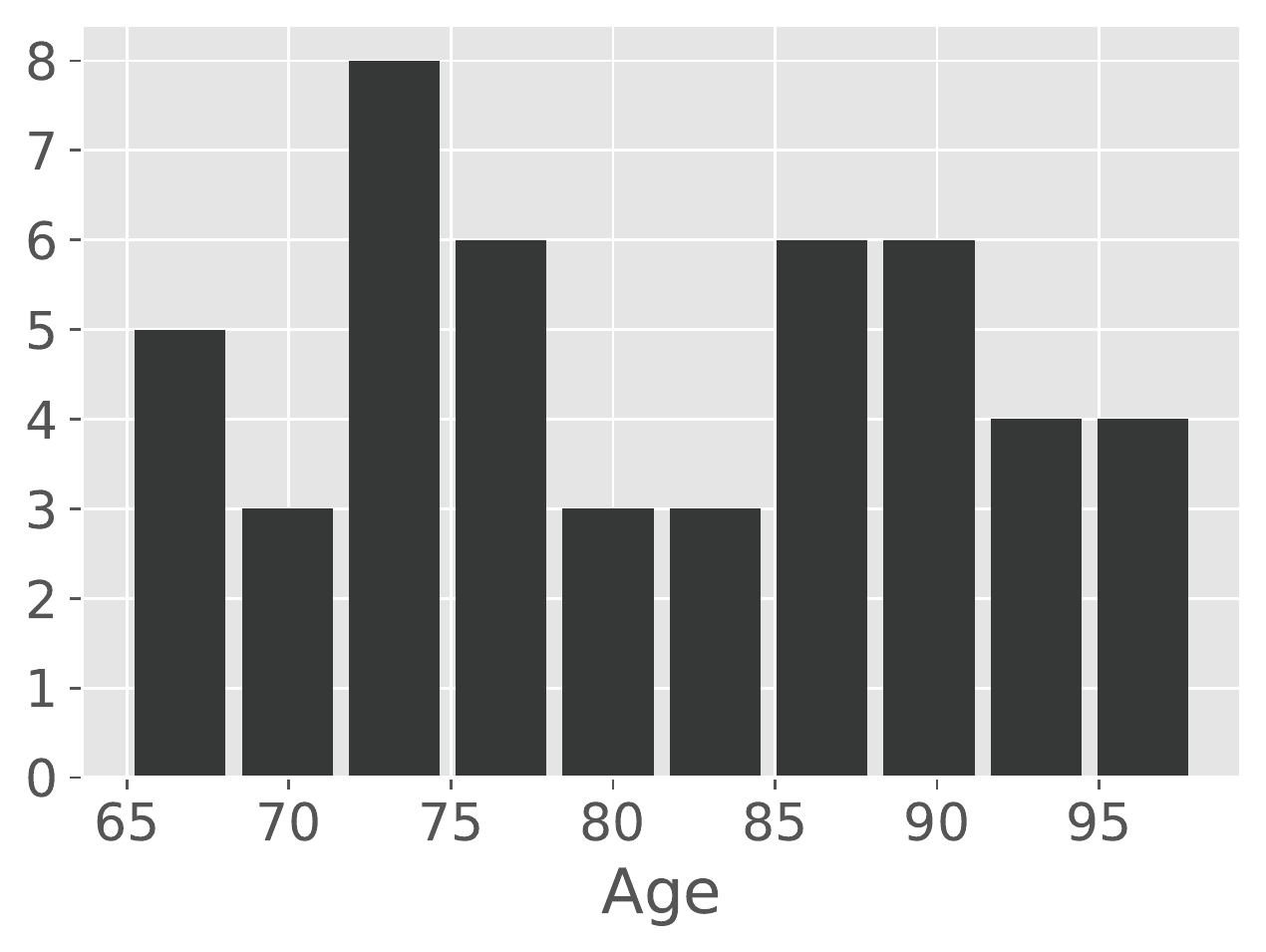}
    \caption{}
    \label{fig:Age_dist}
\end{subfigure}
~
\begin{subfigure}[t]{0.4\textwidth}
    \centering
    \includegraphics[width=\textwidth]{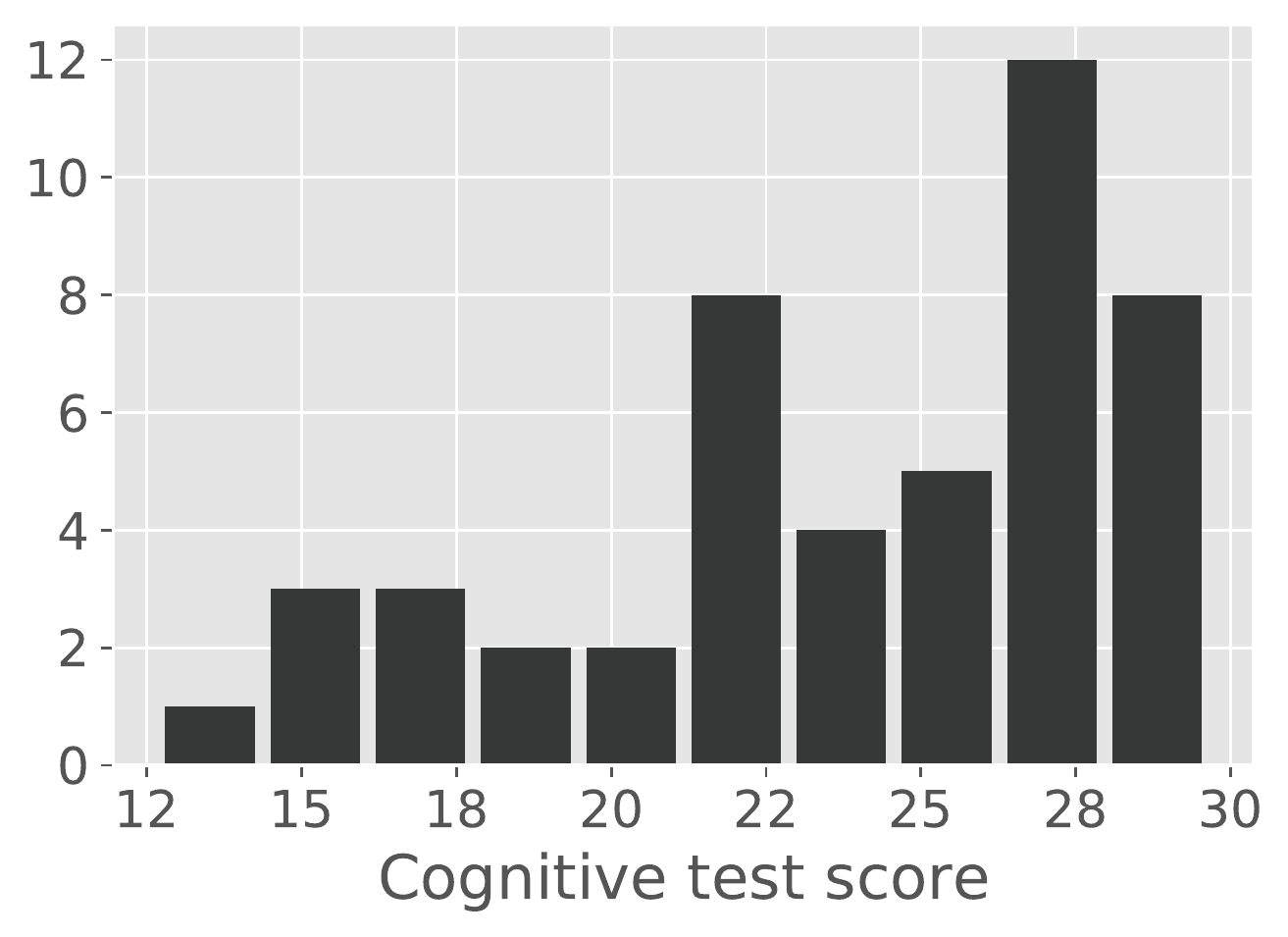}
    \caption{}
    \label{fig:MoCA_dist}
\end{subfigure}
\caption{In \textbf{a)}, the age distribution of the participants is depicted. It follows closely a uniform distribution. In \textbf{b)}, the score of the cognitive test is depicted.}
\end{figure}

In this study, data from a total of 48 people were evaluated (38 women, 13 men).
The participants were all above retirement age with mean age of 81.08 (SD 9.73) and mean cognition score was 23.88 (SD 4.54).
The age distribution is close to uniformly distributed between the ages of 65 and 98.
The Kolmogorov-Smirnov-statistic, when comparing the age values to the uniform distribution is $D=0.064$ with a $p$-value of 0.989.
Both age and cognition distributions are shown in Figure~\ref{fig:MoCA_dist} and~\ref{fig:Age_dist}.
The people were monitored on average for 30.6 days (SD 3.6) - excluding start and end day of measurements.

The partial correlations of the three parameters age, cognition and first reconstruction error were computed.
The results are presented in Table~\ref{tab:correlation}.
There was a slight positive correlation between age and the reconstruction error at $\rho = 0.27$ which was not significant.
Age has a noticeable negative correlation with the score, which was significant at $p < 0.01$.
The reconstruction error and the score had the strongest correlation, at $\rho = -0.42$.
This correlation was highly significant, at $p < 0.005$.

\begin{figure}[H]
\begin{adjustwidth}{-1.5in}{0in}
\begin{flushright}
\centering
\begin{subfigure}[t]{0.4\textwidth}
    \centering
    \includegraphics[width=\textwidth]{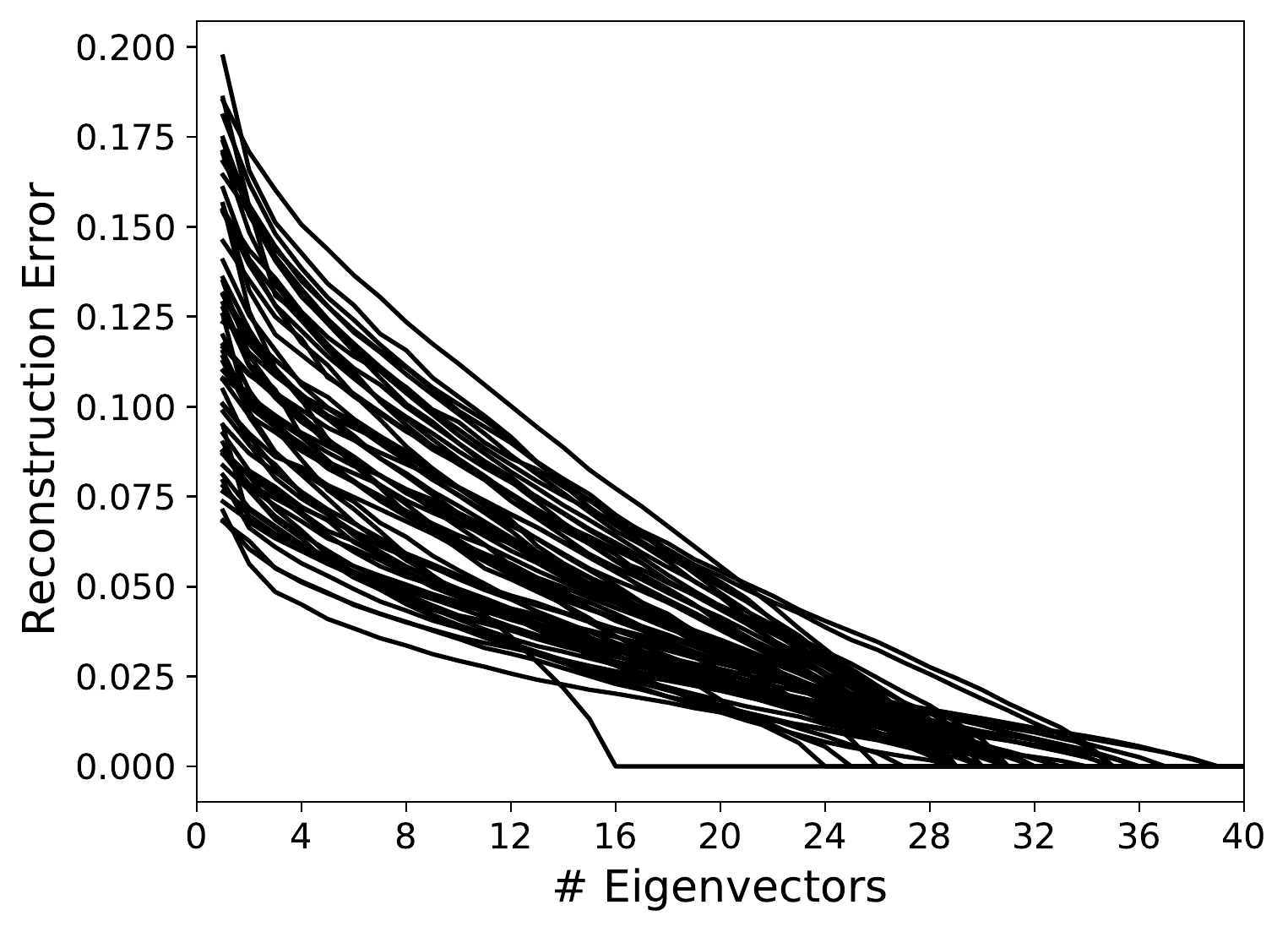}
    \caption{}
    \label{fig:ReconstructionError}
\end{subfigure}
\begin{subfigure}[t]{0.4\textwidth}
    \centering
    \includegraphics[width=\textwidth]{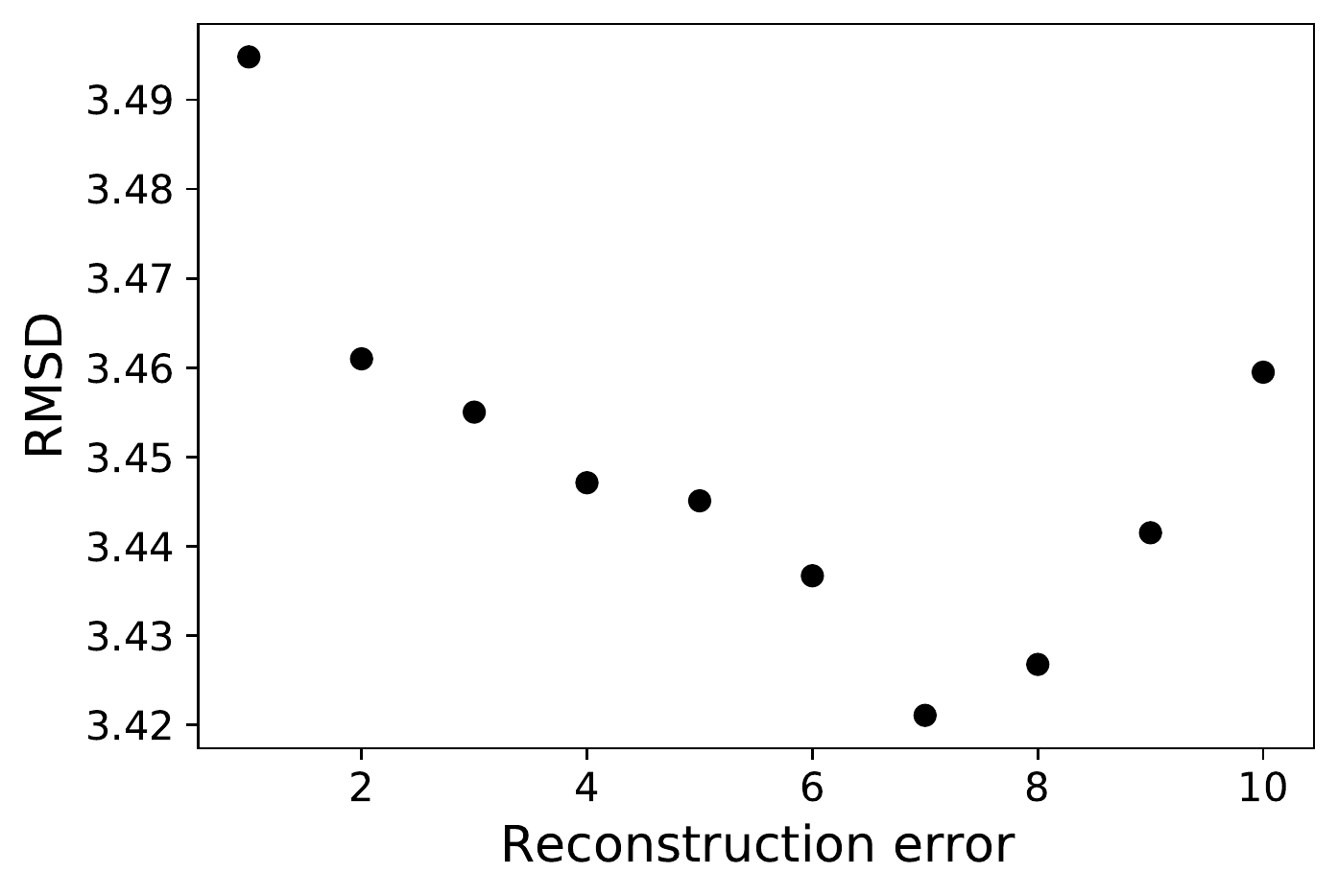}
    \caption{}
    \label{fig:numEigenVec}
\end{subfigure}
\begin{subfigure}[t]{0.4\textwidth}
    \centering
    \includegraphics[width=\textwidth]{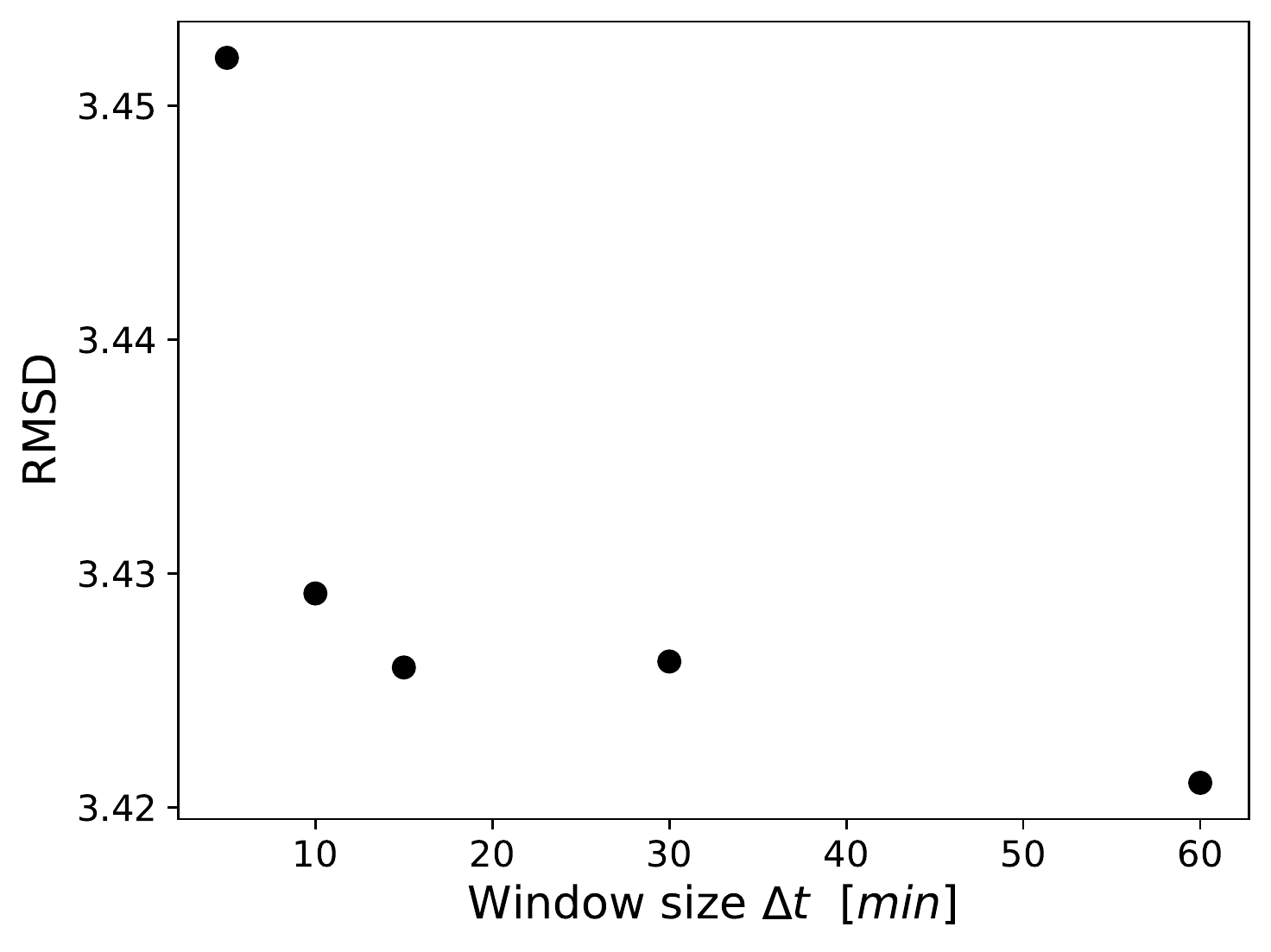}
    \caption{}
    \label{fig:timeres}
\end{subfigure}
\caption{In \textbf{a)}, the normalized reconstruction error for increasing number of used eigenvectors is depicted. In \textbf{b)}, the RMSD of a linear regression for different reconstruction errors is shown. In \textbf{c)}, the optimal window size $\Delta t$ is evaluated. Lowest RMSD is obtained at $\Delta t = 60min \leftrightarrow S = 24$.}
\end{flushright}
\justify
\end{adjustwidth}
\end{figure}

\begin{table}[H]
\caption{Partial correlation of cognitive-score, reconstruction error and age.\newline $^{*}$p-value $< 0.01$; $^{**}$p-value $<0.005$.}
\centering
\begin{tabular}{lc}
\toprule
 	& \textbf{$\rho$}	\\
\midrule
Age vs Reconstruction error & $0.27$\hspace{0.2cm}\\
cognitive ability vs Age		& $-0.38^{*}$	\\
Cognition score vs Reconstruction error		& $-0.42^{**}$\textrm{ }\\
\bottomrule
\end{tabular}
\label{tab:correlation}
\end{table}
Based on the behaviour matrix, the reconstruction errors were computed.
They decreased for increasing number of included eigenvectors up up til their vanishing point when $|D_i|$ eigenvectors were used for the reconstruction.
This is depicted in Figure~\ref{fig:ReconstructionError}, where the segmentation was set at $S=24$ resulting in one hour long time segments.
For other segmentations the structure of the reconstruction errors looked similar.
First, the results of the parameter optimization are presented.
As a baseline, the RMSD of the baseline was computed - the linear regression which was based only on age as a feature and no reconstruction error. 
This resulted in a RMSD of 3.74, higher than any of the regressions including the reconstruction errors.
The RMSD for the cognitive ability prediction was computed for all reconstruction errors and is shown in Figure~\ref{fig:numEigenVec} up to the $10th$ reconstruction error.
The best performance, i.e. lowest RMSD, was obtained when using the $7th$ reconstruction error.
This is shown in Figure~\ref{fig:numEigenVec}, where the segmentation is set at $S=24$.
For all other $S = \{48, 96, 144, 288\}$, similar results were obtained, with the $7th$ reconstruction error being the best choice for the prediction. \newline
In Figure~\ref{fig:timeres}, the RMSD is depicted for $S=\{24, 48, 96, 144, 288\}$ equivalent to window sizes of $\Delta t = \{ 60min, 30min, 15min, 10min, 5min\}$.
The $7th$ reconstruction error was used in this figure.
For $S = 288$, the RMSD is substantially higher than for the other chosen window sizes. 
The results for $S=\{24, 48, 96\}$ are very close together, but the lowest RMSD is obtained for $S=24$, i.e. $\Delta t = 60min$.

\begin{figure}[H]
\begin{adjustwidth}{-1in}{0in}
\begin{flushright}
\centering
\begin{subfigure}[t]{0.3\textheight}
    \centering
    \includegraphics[height=0.2\textheight]{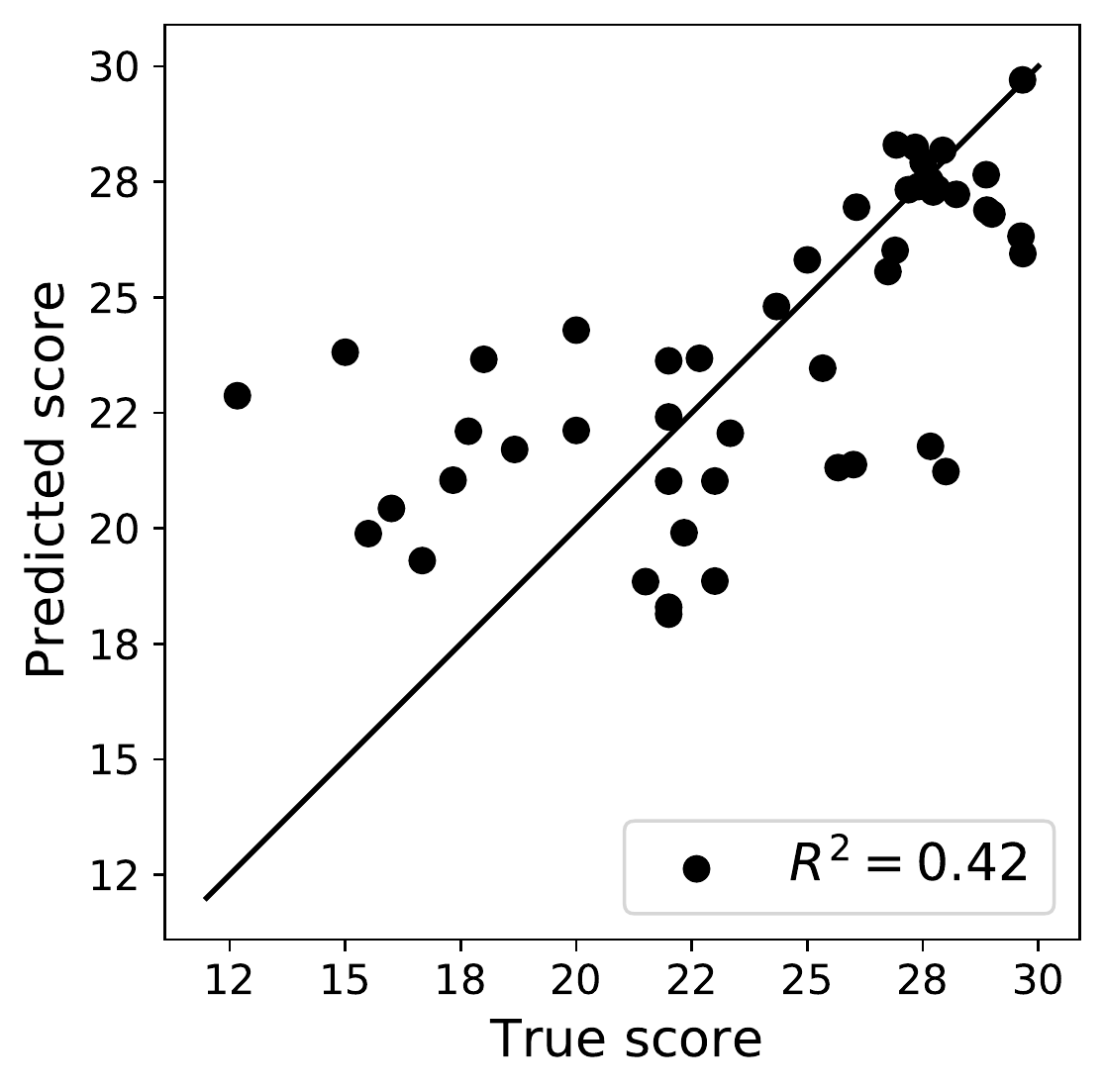}
    \caption{}
    \label{fig:MocaReconstruction}
\end{subfigure}
~
\begin{subfigure}[t]{0.3\textheight}
    \centering
    \includegraphics[height=0.2\textheight]{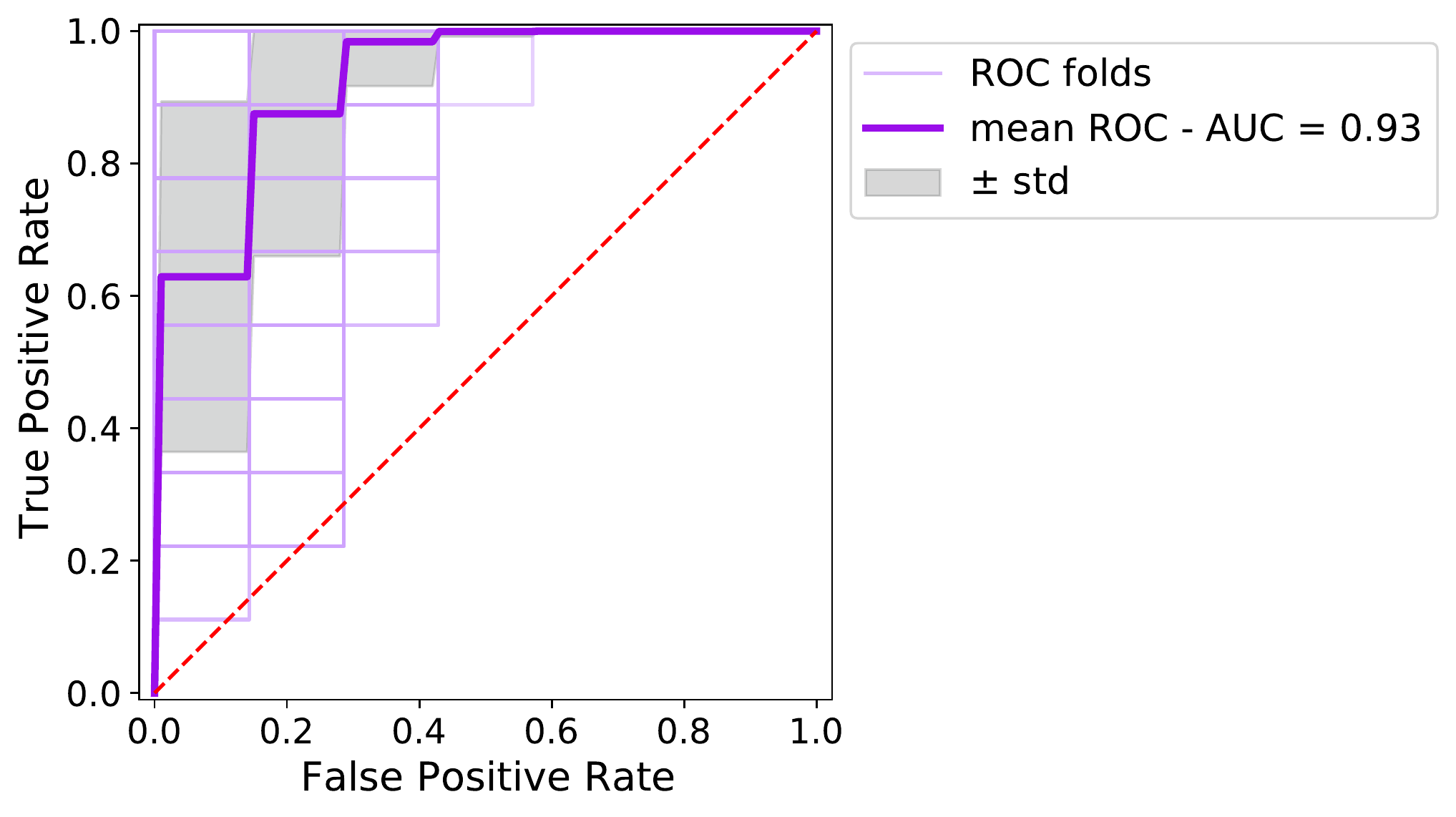}
    \caption{}
    \label{fig:ROC_classes}
\end{subfigure}
\caption{In \textbf{a)} linear regression based on the optimal window size $\Delta t = 60min$ and the optimal $7th$ reconstruction error is shown. 
In \textbf{b)} the violet line is the mean ROC of the classification. The thin lines indicate all individual runs. The average AUC is 0.93}
\end{flushright}
\justify
\end{adjustwidth}
\end{figure}   
The linear regression model was evaluated in a leave-one-out cross-evaluation. 
In Figure~\ref{fig:MocaReconstruction}, the true and predicted scores are depicted together with the coefficient of determination $R^2 = 0.42$.
The black line indicates optimal prediction performance.
The RMSD of this linear regression with window size $\Delta t = 60min$ and using the $7th$ reconstruction error was $RMSD = 3.42$.
For higher scores, the predictions were more concise, while for lower scores the prediction became worse and spread out.

The model performance for the classification task was evaluated and an ROC curve was calculated.
The mean of the ROC, based on a k-fold cross-validation is shown in Figure~\ref{fig:ROC_classes}.
The AUC of the mean ROC was $AUC = 0.93$, but the variation between individual evaluations was high.
The standard deviation of the multiple ROC runs is shown as well.

\section*{Discussion}
In this work, we have shown, how PIR-sensor based location information could be used to gain insights into the cognitive ability of the older individuals monitored over one month.
Based on the location information, an eigendecomposition was made which is sensitive to the regularity in the behaviour patterns.
The more predictable and regular the participants moved around in their apartment, the fewer eigenvectors were required to reconstruct their behaviour.
Lower cognitive abilities has been found to be associated with a loss of routine~\cite{urwyler_cognitive_2017}. 
A loss of routine, or more erratic behaviour, is harder to map onto fewer eigenvectors.
The reconstruction error will be larger than compared to regular and predictable behaviour.

In our evaluation, we have looked at two usages of the reconstruction error in order to predict the cognitive ability.
First, we conducted a prediction of the cognition score, based on the reconstruction error.
Second, we did a classification into a healthy group versus a group with mild to sever cognitive impairment.

For both the prediction as well as the classification of the score, two parameters were optimized, the window-length $S$ and the choice of reconstruction error.
The error for the time window of length 60 minutes is the smallest, but only by a small margin as compared to the other time windows of length 10 minutes up to 30 minutes.
For the time window of length 5 minutes ($S = 288$), the error increases substantially.
It is likely that in our everyday routine, there is a lower boundary on our time precision.
A boundary, under which it is no longer possible to distinguish between routine behaviour and erratic or chaotic behaviour.
For example, if we set an alarm for getting up in the morning, the time we actually get up might still differ by a few minutes, influenced by our mood, our sleep quality or something else.
By checking different window lengths, it seems this time window is between five and ten minutes.
This would mean, that in our routine behaviour, we tend to be exact down to a lower resolution of 10 minutes. 
Another consideration is the computational time.
The computation of the eigenvalues is considerably more demanding for a $|K|\cdot S \times |K|\cdot S = 5\cdot 144\times 5\cdot 144$ matrix than for a $5\cdot 24 \times 5\cdot 24$ matrix, and thus the choice of time resolution should take the computational resources into consideration.

The other parameter that was optimized was the choice of reconstruction error.
In the work of Eagle et al.~\cite{eagle_eigenbehaviors:_2009}, the number of eigenvectors needed to achieve a certain level of reconstruction was used to distinguish between different population groups.
In a similar matter, we looked for the best number of eigenvectors needed for the reconstruction error being able to best distinguish interindividual differences in cognition.
The most common every day structural routines are covered by the first few eigenvectors.
Due to repeating structures of different time frames - hourly, daily and weekly - too few eigenvectors would not be able to cover all of this behaviour.
On the other hand, when adding too many eigenvectors for the reconstruction, they no longer explain predictable behaviour but actual behavioural noise.
This behavioural noise is probably best explained by our own timely inaccuracies as discussed in the previous paragraph as well as disturbances from the outside world.
Interestingly, the optimal number of eigenvectors found in our analysis was always seven. 
While this could be coincidental, it could just as well hint to the periodicity of the weekly behaviour patterns.
A similar discovery was made by~\cite{eagle_eigenbehaviors:_2009}, where certain eigenvectors cover specific behavioural aspects, such as weekends or breaks. 

For the subgroup classification, the data was split into two groups.
The group with a cognition score at or above 26, and the group with a cognitive ability below 26.
The rationale behind this split was the close relationship to the cognitive ability.
A cognitive ability at or above 26 is commonly considered to coincide with normal cognitive ability, whereas a cognitive ability below 26 is connected to MCI or AD~\cite{nasreddine_montreal_2005}.
The resulting AUC is surprisingly large. 
But considering the comparably small set, 32 data points in the training set and 16 data points in the test set, this value should be taken with a grain of salt.
There is still room for variance, and the evaluations would best be repeated with larger data sets.
Due to the small sample size, further splitting of the data into a third group with cognitive ability below 17, as suggested for AD~\cite{nasreddine_montreal_2005} could not reasonably be performed.

Our method shows good prediction behaviour for higher cognition scores, but worse performance when the actual cognitive ability score is below 20.
On one hand, we have fewer data points in that area to train a model with, which could explain this lower performance. 
On the other hand, there are numerous different reasons for low cognitive ability score; reduced language comprehension, working memory, concentration and attention are some of the abilities needed to reach high scores.
As the evaluation does not differentiate between the different causes for lower scores, their effect on the movement behaviour is variable.
This is not taken into consideration in this evaluation.
In a future study, more thorough evaluation of the participants and classification of their cognitive ability could improve on these results.

While not all causes for a lower cognitive ability score might lead to a change in movement patterns, there might also be factors present causing changing patterns, that are not represented through cognitive abilities. 
An example was given by Paraschiv-Ionescu et al. in their study covering chronic pain and its effect on physical activity patterns~\cite{paraschiv-ionescu_barcoding_2012}.
Furthermore, there is a reasonable chance, that some causes for lower a cognitive ability score might even favor the regularity of patterns and increase them.
These uncertainties indicate the limitations of this method.


In our study, we have used participants' chronological age as an additional feature to improve our predictions. But we have only looked at people above the age of 65 years.
We do not expect this method to be directly applicable to a younger population.
As most younger adults are likely to have a cognitive ability around 30, a saturation effect is expected to kick in, making the usage of linear regression as a model no longer a good choice.
Nevertheless, the usage of the reconstruction error might still be a valuable feature for other models.

The data we have used in this study covers around four weeks of monitoring.
The cognitive ability of people is not expected to change within 
It would be interesting to assess, whether longer monitoring periods could either improve the prediction of cognition, or alternatively be used to monitor change in the cognitive abilities.

\section*{Funding}
This research was funded by the Hasler Foundation grant number 16072.

\section*{Ethics}
The study was conducted according to the guidelines of the Declaration of Helsinki, and approved by Ethics Committee of the Canton of Bern, Switzerland (KEK-Nr. 406/16).

\section*{Declaration of Interest}
The authors declare no conflict of interest.
The funders had no role in the design of the study; in the collection, analyses, or interpretation of data; in the writing of the manuscript, or in the decision to publish the~results.

\newpage
\begin{adjustwidth}{-1.5in}{0in}
\bibliography{main}

\bibliographystyle{abbrv}

\end{adjustwidth}

\end{document}